\documentclass[a4paper,fleqn]{cas-dc}

\usepackage[authoryear,longnamesfirst]{natbib}

\def\tsc#1{\csdef{#1}{\textsc{\lowercase{#1}}\xspace}}
\tsc{WGM}
\tsc{QE}
\tsc{EP}
\tsc{PMS}
\tsc{BEC}
\tsc{DE}

\usepackage{comment}
\usepackage{tabularx,booktabs}
\usepackage{lipsum}

\usepackage[pscoord]{eso-pic}

\newcommand{\placetextbox}[4]{
  \setbox0=\hbox{#4}
  \AddToShipoutPictureFG*{
    \if#3r
    \put(\LenToUnit{\paperwidth-#1},\LenToUnit{\paperheight-#2}){\vtop{{\null}\makebox[0pt][r]{\begin{tabular}{r}#4\end{tabular}}}}%
    \else
    \put(\LenToUnit{#1},\LenToUnit{\paperheight-#2}){\vtop{{\null}\makebox[0pt][l]{\begin{tabular}{l}#4\end{tabular}}}}%
    \fi
  }%
}%

\begin{document}
\let\WriteBookmarks\relax
\def\floatpagepagefraction{1}
\def\textpagefraction{.001}

\shorttitle{Self-Supervised vs Supervised Training for Organoids Segmentation}    

\shortauthors{Asmaa Haja et al.}

\title [mode = title]{Self-Supervised Versus Supervised Training for Segmentation of Organoid Images}  

\tnotemark[1,2]

%

\author[1]{Asmaa Haja }[
    orcid= 0000-0002-4116-2167]


\cormark[1]


\ead{a.haja@rug.nl}

\credit{
 Methodology, 
 Software, 
 Resources,
 Writing - final draft, 
 Data creation, 
 Investigation,
 Supervision
}

\affiliation[1]{organization={Bernoulli Institute},
            addressline={University of Groningen}, 
            country={The Netherlands}}

\affiliation[2]{
            addressline={University of Groningen}, 
            country={The Netherlands}}

\author[2] {Eric Brouwer }
\credit{
 Conceptualization, 
 Methodology, 
 Writing - original draft, 
 Data creation, 
 Investigation, 
 Visualization
}
\ead{e.b.brouwer@student.rug.nl}

\author[1] {Lambert Schomaker }[orcid= 0000-0003-2351-930X]
\credit{
 Conceptualization, 
 Methodology, 
 Supervision, 
 Project administration,
 Writing - review \& editing.
}

\ead{l.r.b.schomaker@rug.nl}



\cortext[1]{Corresponding author}

\begin{abstract}
The process of annotating relevant data in the field of digital microscopy can be both time-consuming and especially expensive due to the required technical skills and human-expert knowledge. Consequently, large amounts of microscopic image data sets remain unlabeled, preventing their effective exploitation using deep-learning algorithms. In recent years it has been shown that a lot of relevant information can be drawn from unlabeled data. Self-supervised learning (SSL) is a promising solution based on learning intrinsic features under a pretext task that is similar to the main task without requiring labels. The trained result is transferred to the main task - image segmentation in our case. A ResNet50 U-Net was first trained to restore images of liver progenitor organoids from augmented images using the Structural Similarity Index Metric (SSIM), alone, and using SSIM combined with L1 loss. Both the encoder and decoder were trained in tandem. The weights were transferred to another U-Net model designed for segmentation with frozen encoder weights, using Binary Cross Entropy, Dice, and Intersection over Union (IoU) losses. For comparison, we used the same U-Net architecture to train two supervised models, one utilizing the ResNet50 encoder as well as a simple CNN. Results showed that self-supervised learning models using a 25\% pixel drop or image blurring augmentation performed better than the other augmentation techniques using the IoU loss. When trained on only 114 images for the main task, the self-supervised learning approach outperforms the supervised method achieving an F1-score of 0.85, with higher stability, in contrast to an F1=0.78 scored by the supervised method. Furthermore, when trained with larger data sets (1,000 images), self-supervised learning is still able to perform better, achieving an F1-score of 0.92, contrasting to a score of 0.85 for the supervised method.
\end{abstract}


\begin{highlights}
\item Organoid images
\item High-throughput
\item Detection
\item Segmentation
\item Self-supervised
\item Microscopic images analysis
\end{highlights}


\maketitle
\placetextbox{0.1cm}{0.5cm}{r}{This article was submitted to Elsevier CMPBUP \\ 
Date of submission: 20th September 2022 
}

\section{Introduction\label{sec:intro}}

With the advances in high-throughput imaging technology, it is currently possible to produce a large number of microscopic images in a short period of time \cite{haja2021fully}. 
Comprehensive analyses of biological images are required for medical diagnosis, and disease comprehension \cite{zhang2020discriminative}. 
Detecting diseases through manually analyzing the rich biological information in microscopic images is challenging since it is time-consuming, it demands domain knowledge in the field, is biased to the individual human experts and therefore not entirely accurate. Because manual analysis is an exhausting task it can lead to fatigue leading to human errors \cite{adhikari2021sample} \cite{zhu2021hard}.
Accordingly, research in the biological field can be a slow and arduous endeavour. Methods from the field of deep learning may have the potential to automatically extract relevant information from biological images.

Deep learning has recently found major successes in the automation of data processing, data manipulation, and understanding of data \cite{dargan2020survey}. In regards to biomedical image processing, deep networks have been able to demonstrate exceptional performance in tasks such as classification \cite{mai2022online}, detection, and segmentation \cite{vu2019methods} through \textbf{supervised} learning. Their performance and success rely heavily on the use of labelled or annotated data \cite{sharma2021comprehensive}. The process of annotating relevant data in this context still involves manual labour, leaving researchers with a similar issue where large amounts of biomedical image data sets remain unlabeled, keeping this valuable material nearly useless for their intended tasks. In contrast, large sets of intrinsic information can be drawn from the data even if it remains unlabeled \cite{jaiswal2020survey}. \textbf{Self-supervised learning} (SSL) is a promising new approach for processing and extracting relevant information from data sets consisting of a higher proportion of unlabeled images in comparison to the annotated images \cite{chen2019self} \cite{azizi2021big}. 

In self-supervised learning, an artificial 'pretext' task that does not require manual labeling is used to train a neural network, that is used to perform the main intended task in a second stage. Because the pretext task is completely based on the given data only, the associated training set can be very large, even much larger than would be possible for an annotated data set.

Deep learning models using the SSL paradigm allow for models to familiarize themselves with the data of interest, where the acquired 'knowledge' is transferred to a secondary supervised approach that will only need a limited effort in training afterwards.

The intention of this work is to employ the SSL paradigm for biomedical imaging, specifically on organoid-culture images. In essence, organoids are self-organizing three-dimensional structures grown from \textit{in vitro} stem cells, with the ability of mimicking its \textit{in vivo} tissue counterpart \cite{de2018organoids}. This ability can be used as a powerful medical tool. For instance, it can be used to indicate different diseases based on changes in their morphology (shape, and structure) \cite{kretzschmar2021cancer}.
Precise measurements of organoids' morphology can be achieved by segmenting organoid objects in the image dataset.

With this aim in mind, this paper explores the ability to use the SSL method to segment organoid culture images, as well as to compare the supervised with the SSL approach in order to observe the amount of data that is required to develop a robust model. 
To the authors’ knowledge, no work exists in the literature that explores the detection and segmentation of organoid images using the self-supervised concept, which also shows the novelty of this work. 

The main contributions of this work are summarized as follows:
\begin{itemize}
    \item Studying and analyzing the SSL approach on organoid data;
    \item Investigating the performance of various augmentation techniques on the SSL main task;
    \item Evaluating the performance of SSL on the main task over a range of training-set sizes;
    \item Comparing SSL versus traditional supervised training;
    \item Evaluating and comparing different loss functions for the pretext task as well as the main task for both SSL and supervised approaches;
    \item Investigating the effect of freezing the encoder stage in a U-Net architecture.
\end{itemize}

This work is organized into six sections, with the following structure: Section \ref{sec:related_work} presents a review of the related works providing a deeper insight into organoids research, semantic segmentation, supervised learning, and self-supervised learning. Section \ref{sec:method} describes the method of the investigation where the organoids data, loss functions, and the supervised as well as the SSL frameworks are discussed. Section \ref{sec:design} is reserved for the experimental design, describing the data distribution and the implementation details. Section \ref{sec:results} presents a discussion of the results, and lastly section \ref{sec:conclusions} concludes this work and presents possible future studies.
\section{Related Works\label{sec:related_work}
}

\subsection{Organoids research\label{sec:organoids_research}
}
\textit{In vivo} tissues and organs studies  conducted on animal and especially on humans can be hampered by the expensive costs, limited resources
, and ethical considerations \cite{rossi2018progress}. 
The developed of \textit{in vitro} stem cell culture opened the door for a new area of research since it overcomes these issues \cite{graudejus2018bridging}.
Organoids are self-organizing three-dimensional structures grown from \textit{in vitro} stem cells that  mimics its counterpart \textit{in vivo} organ
\cite{tuveson2019cancer} \cite{de2018organoids}, \cite{kratochvil2019engineered} \cite{corro2020brief}. 
Organoids are essential due to their large scale applications in different domains, 
includes but not limited to 
modeling organ development and disease \cite{rossi2018progress}, cancer \cite{drost2018organoids}, personalized medicine and drug discovery \cite{wang2021ms}, and regenerative medicine \cite{marchini2022synthetic}.

The organoid growth and its morphological changes play an important role in developing appropriate drugs since these changes affect organoids' responses to treatments  \cite{karolak2019morphophenotypic} \cite{hoang2021engineering}. In other words, an improper estimate of the organoid's area or volume results in an incorrect measurement of the organoid's responsiveness to treatments \cite{karolak2019morphophenotypic}. Therefore, the need to accurately measure organoids' morphology is crucial. 
Towards achieving precise morphological measurements of organoids' morphology and growth, in this work, models from deep learning field are employed.

\subsection{Semantic Segmentation\label{sec:Semantic_segmentation}
}
Semantic segmentation technique is one of the deep learning approaches to measuring the morphological characteristics of the objects (in this case organoids). 
The goal of the semantic segmentation task is to assign a class label to every pixel in the image. 
Groups of neighbouring pixels belonging to the same class label are semantically considered to describe the same object.
A deep learning model aiming at semantically segmenting an input image would output an image with the same shape, in which pixel values range between 0 and the number of class labels. In this work, we focus on segmenting one class, e.g the organoid class. Hence, the output image contains only two values 0 and 1 describing background / no-object and object, respectively.

Multiple types of software already exist that attempt to segment organoid culture images. One example is the software package OrganoSeg \citep{borten2018automated}, which provides an intuitive, graphical user interface for quantifying transmitted-light images of 3D spheroid and organoid cultures. However, this software requires some manual work from the user to define and finetune thresholds and parameters used for separating the foreground from the background. 
Another example is the OrganoID \citep{matthews2022organoid}, a robust image analysis platform that automatically recognizes, labels, and tracks single organoids, pixel-by-pixel, in brightfield and phase-contrast microscopy experiments using deep learning. This software employs Sobel operators, Gaussian filter, and watershed for detecting single organoid. These techniques are highly affected by the image quality (e.g. change in brightness) and cannot be generalized to all microscopic data.
Another example is the deepOrganoid model, which is based on a deep learning technique that can be used as a fully automatable analytical tool for high-throughput screens that rely on organoid cultures \citep{powell2022deeporganoid}. Although researchers in the organoid field can utilize this model by retraining it on their dataset, the problem of possessing sufficiently large labelled data beforehand to train the model is still an issue in this field.
All these tools cannot be generalized for various organoid datasets with a  limited amount of annotation as they are trained using supervised learning.


\subsection{Self-supervised Learning\label{subsec:Self-supervised_Learning}}
There exist numerous techniques for teaching a model to learn a particular task on a given dataset. A well-known technique is the \textbf{supervised learning}, which can be described as a 'supervisor' who instructs the learning system on how to learn the data by mapping the annotated data to their associate predefined labels \cite{cunningham2008supervised}.
Unfortunately, deep learning models based on a supervised approach require a huge number of data, particularly annotated data, to learn from. 
This is not the case in the biology and medicine fields. 
Producing and annotating data are both time-consuming tasks that also require professional knowledge \cite{yu2016automatic}.

\textbf{Self-supervised learning (SSL)} addresses this issues by extracting useful representations from large-scale unlabeled data without any manual annotations \cite{wang2022self}. 
This can be realised by splitting the goal into two tasks: pretext task and main task. 

\textbf{Pretext task}
The idea of the pretext task is based on the fact that a convolutional neural network extracts different levels of information from its different layers. The shallow layers of CNN capture the low-level features (e.g., texture and gradient), whereas deeper layers capture the high-level features (e.g., semantic information) of the image \cite{li2017compactness}
\cite{chen2021ssl++}
\cite{zhou2021siamese}. 
Therefore, the pretext task utilizes this fact by applying a transformation to the unlabelled input data and then requiring the model to predict the properties of the transformation from the transformed data \cite{misra2020self}.
Prediction the rotation of an images \cite{gidaris2018unsupervised}, solving jigsaw puzzle \cite{noroozi2016unsupervised}, and finding a relative position between two random patches from one image \cite{doersch2015unsupervised} are some pretext tasks examples.
As a consequence, the model learns the low- and middle-level features on the unlabeled data and can then be fine-tuned on the limited amount of labelled data. 

\textbf{Main task} The ultimate task, which is segmenting organoids in this case, retrains the trained model from the pretext task in order to capture the high-level features. Since the low- and middle-level features are almost captured in the model weights, the model only needs limited labelled data to learn the main task.

Briefly, the self-supervised approach is based on pretraining a trained model on large unlabeled (transformed) data and then transferring the model’s knowledge to the main task \cite{zaiem2021pretext}. 
To put it differently, the self-supervised training removes the dependency on manual labels \cite{zhang2021depa}.



\begin{figure*}[htb!]
\centering
\includegraphics[width=1.0\textwidth]{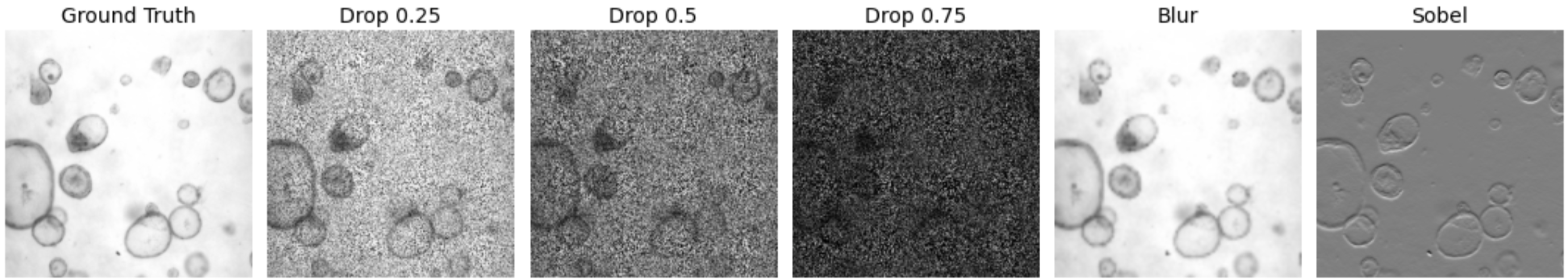}
    \caption{ \textbf{Three augmentation techniques.} A random images was selected. From left to right: the ground truth images, 25\% pixel drop, 50\% pixel drop, 100\% pixel drop, Gaussian Blurring and Sobel Filtering.
    \label{FIG:augmented_data}}
\end{figure*}

\section{Method\label{sec:method}}

\subsection{Data Augmentation\label{subsec:data}}

\begin{figure}[ht!]
\centering
\includegraphics[width=0.45\textwidth]{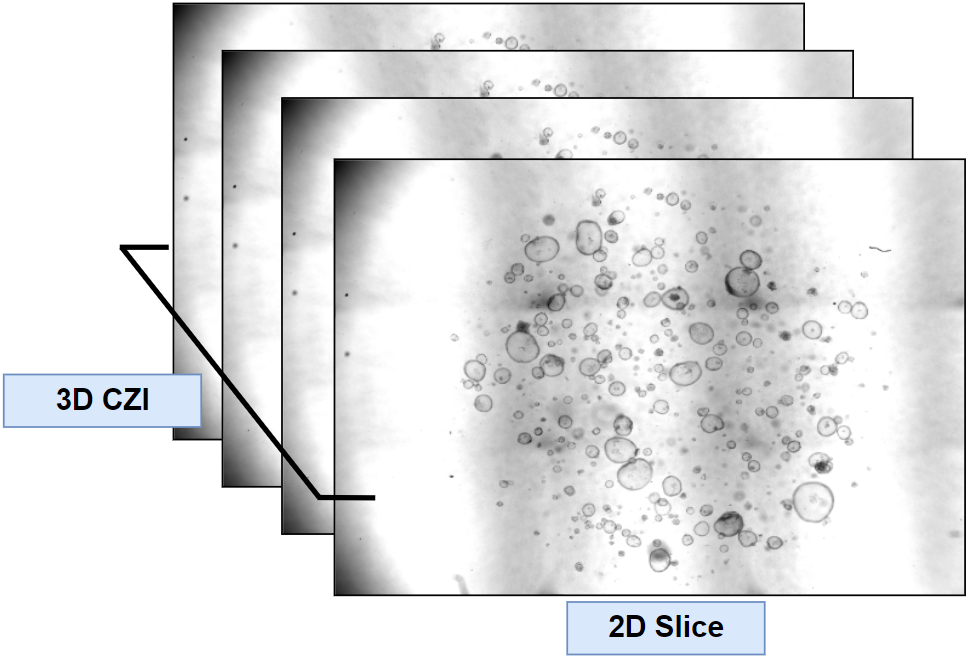}
    \caption {An example of a CZI: A 3D image made up of 2D slices, called stacks, at different depths of the organoid culture.}
    \label{fig:czi_image}
\end{figure}

The approach is evaluated on liver progenitor organoids images provided by the University Medical Center Groningen. These CZI\footnote{CZI file consists of a number of stacks, each captured at different depth of the 3D organoid culture} images (Figure \ref{fig:czi_image}) were captured at 5 different time points (0 to 96 hours with an interval of 24 hours) in two growing conditions: (1) liver progenitor organoids have been grown in either complete medium, or (ii) in medium lacking all amino, are essential for growth \cite{haja2023towards}. A total of 10 CZI images, each consisting of 14 stacks, are studied, yet, an average of 4 stacks per image were evaluated as the lower and higher stacks do not contain many information. The size of each stack image is 3828x2870 pixel. 

Using the OrganelX\footnote{https://organelx.hpc.rug.nl/organoid/} service, the corresponding true mask images were created, in which the pixel values 1 and 0 describe object or no-object, respectively. A manual correction took a place to improve the segmentation on some organoids' border using the system.
The original stacks were also cropped into smaller images of 636x636 pixels (called crops) using a sliding window technique increments every 60 pixels, and then resized to 320x320 pixels in order to decrease the training time. Crops containing less than 5\% objects were not considered. 
The total number of created crops is around 25.000.

To perform the pretext task, explained in sections \ref{subsec:Self-supervised_Learning} and \ref{subsec:SSL_framework}, three augmentation techniques were performed on the images: \textbf{(1)} Pixel drop: random noise is added to an image by randomly dropping pixels from the image, \textbf{(2)} Gaussian blurring: image resolution is cut in half by performing a Gaussian blur function, \textbf{(3)} Sobel filtering: a Sobel operation is applied on the image resulting in an emphasis on object edges. 
Image rotation of 0$^{\circ}$, 90$^{\circ}$, 180$^{\circ}$ and 270$^{\circ}$ was also used as an augmentation technique to increase the number of total images, resulting in around 100.000 cropped and augmented images being used for each augmentation technique. Figure \ref{FIG:augmented_data} displays a randomly selected image from the dataset with all augmentation techniques applied. 

\subsection{Loss Functions\label{subsec:loss_functions}}

Sections \ref{sec:ssim_loss} and \ref{sec:ssim_l1_loss} describe loss functions used for the pretext task, whilst sections \ref{sec:bce}, \ref{sec:dice}, and \ref{sec:iou} describe loss functions used in the main task.

\subsubsection{SSIM}\label{sec:ssim_loss}
The Structural Similarity Index Metric (SSIM), proposed by \cite{wang2004image}, measures the similarity between two given images. An image is divided into various windows, where x and y indicate their respective window of the two images with shared sizes $N \times N$. The score is calculated as shown in equation \ref{eq:ssim}.

\begin{equation}\label{eq:ssim}
    SSIM(x,y) = \frac{(2\mu_x\mu_y+c_1)(2\sigma_{xy}+c_2)}{(\mu^{2}_{x}+\mu^{2}_{y}+c_1)(\sigma^{2}_{x}+\sigma^{2}_{y}+c_2)}
\end{equation}

With:
$\mu_{x}$ the average of $x$;
$\mu_{y}$ the average of $y$;
$\sigma^{2}_{x}$ the variance of $x$;
$\sigma^{2}_{y}$ the variance of $y$;
$\sigma_{xy}$ the covariance of $x$ and $y$;
$c_1$ and $c_2$ as constants to stabilise cases with weak denominators (i.e. zero). The SSIM value can then be used to compute the SSIM loss as shown in equation \ref{eq:ssim_loss}.

\begin{equation}\label{eq:ssim_loss}
    L_{SSIM}(x,y) = 1 - SSIM(x,y)
\end{equation}

\subsubsection{SSIM-L1}\label{sec:ssim_l1_loss}
In some cases, the SSIM loss suffers from sensitivity biases. Resulting from this, when images are restored, changes in colour or brightness can be observed \cite{zhao2015loss}. In contrast, the mean absolute error, also known as the L1 loss (shown in eq. \ref{eq:mae}), suppresses this factor more heavily. Here, $n$ indicates the number of pixels, $Y$ the target output, and $\hat{Y}$ the model's predicted output.

\begin{equation}\label{eq:mae}
    L_{MAE} = 1 - \frac{1}{n}\sum^n_{i=1}|Y_i - \hat{Y}_i|
\end{equation}

In principle, to get clearer image restorations, the L1 loss is combined with the SSIM loss function in a symmetrical manner, and is shown in eq. \ref{eq:ssim_l1}.

\begin{equation}\label{eq:ssim_l1}
    L_{SSIM-L1} = \frac{1}{2}L_{MAE} + \frac{1}{2}L_{SSIM}
\end{equation}

\subsubsection{Binary Cross Entropy}\label{sec:bce}
Binary cross entropy (BCE) \cite{jadon2020survey} compares the probability of the model's predicted output class $\hat{Y}$ to the actual class label $Y$ within a range of 0 and 1, as shown in equation \ref{eq:bce}. Due to this nature, it can then be used for the task of binary pixel-wise classification necessary for the segmentation task, as mentioned in section \ref{sec:Semantic_segmentation}. In this case, $n$ refers to the number of pixels present in the image.

\begin{equation}\label{eq:bce}
    L_{BCE}(Y, \hat{Y}) = -\frac{1}{n} \sum^{n}_{i=1}(Y_{i} \cdot log \hat{Y}_{i} + (1-Y_i) \cdot log(1-\hat{Y}_i))
\end{equation}

Generally, with cross-entropy functions, the gradients with respect to the logits produce smoother loss values allowing for better stability in training when compared to other loss functions in the same domain \cite{jadon2020survey}.

\subsubsection{Dice}\label{sec:dice}
Dice loss is a commonly used loss function in the context of semantic segmentation \cite{yeung2022unified}. The similarity between the output image $\hat{Y}$ and target output $Y$ is computed as shown in equation \ref{eq:dice}. Here, $\epsilon$ denotes a constant value for cases with a weak denominator, known as a smoothing factor \cite{li2019dice}.

\begin{equation}\label{eq:dice}
    L_{dice}(Y, \hat{Y}) = 1 - \frac{2 \cdot \sum Y \cdot \hat{Y}}{\sum Y^2 + \sum \hat{Y}^2 + \epsilon}
\end{equation}

In contrast to cross-entropy functions, the dice metric can cause gradients to blow up to large numbers, often resulting in unstable training \cite{eelbode2020optimization}. However, dice losses are more robust when presented with imbalanced datasets, which is relatively common in semantic segmentation; typically, the background accounts for a more significant portion of the pixels than the object of interest. This is also the case for organoid images.

\subsubsection{Jaccard}\label{sec:iou}
The Jaccard loss, also referred to as the Intersection over Union (IoU), is less commonly used than the dice loss but is also a powerful tool for semantic segmentation \cite{bertels2019optimizing}. Here, the sum of the product between the predicted output $\hat{Y}$ and target output $Y$ is computed, then divided by its union as shown in equation \ref{eq:iou}. Again, $\epsilon$ is used as a constant to prevent a zero division.

\begin{equation}\label{eq:iou}
    L_{IoU}(Y, \hat{Y}) = 1 - \frac{\sum(Y\cdot \hat{Y})}{\sum(Y + \hat{Y}) - \sum(Y \cdot \hat{Y}) + \epsilon}
\end{equation}

The Jaccard loss suffers from a similar problem to the dice loss regarding blowing up gradients. However, like the dice loss, it works well with an imbalanced dataset with the addition of scale invariance, granting relevance to smaller objects \cite{bertels2019optimizing}. The ability to include such smaller objects is pertinent to the organoids data set as images typically have both large and small organoids spread across the image.

\subsection{Model Structures\label{subsec:model_structure}}

\begin{figure}[htb!]
\centering
\includegraphics[width=0.5\textwidth]{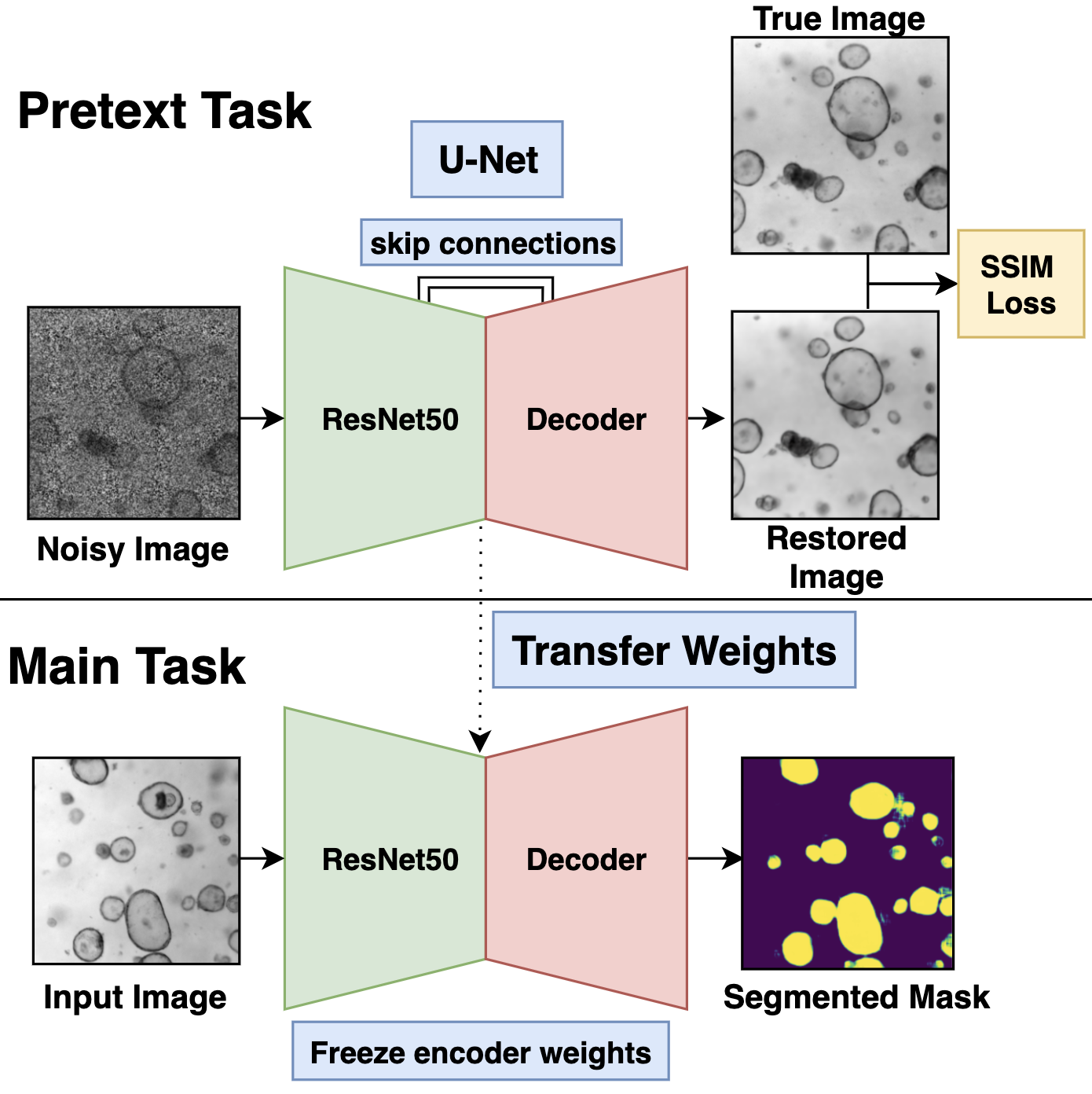}
    \caption{ \textbf{Self-supervised pipeline for organoid segmentation.} \textbf{Top:} Pretext task - The U-Net model consists of a ResNet50 encoder and a decoder connected via skip connection. The ResNet50 is trained to restore augmented images to their original form. The restored image is compared to the ground truth image, and the SSIM or SSIM-L1 loss is computed. \textbf{Bottom:} Main task - The same network as in the pretext task, yet, with a frozen encoder. The decoder learns to segment the ground truth images. 
    \label{FIG:ssim_pipeline}}
\end{figure}

\subsubsection{Self-Supervised Framework\label{subsec:SSL_framework}}

The self-supervised framework consists of two phases, as demonstrated in Figure \ref{FIG:ssim_pipeline}. A pretext task is essentially designed to push the model to learn the semantic features of the input images; by allowing the model to 'pre-train' on the data, the model is able to familiarize itself with the data. The proposed pretext task is to perform image restoration on augmented images (section \ref{subsec:data}). For this task, a U-Net with a ResNet50 encoder as the backbone paired with the decoder are trained in tandem to restore an augmented input image. The data is augmented through three techniques as described in section \ref{subsec:data}. The output of the U-Net is compared to the ground truth image where the SSIM and SSIM-L1 losses are computed as described in section \ref{subsec:loss_functions}. The concept for this task is to train the network to attempt to generate organoid features, which would set up the network's weights for the second phase. In the second phase, the learnt weights are transferred to perform the main task of segmentation on the images of organoids, where the encoder weights are frozen and only the decoder is re-trained using the loss functions described in sections \ref{sec:bce}, \ref{sec:dice}, and \ref{sec:iou}.

\subsubsection{Supervised Framework\label{subsec:SPV_framework}}

The supervised framework consists of two approaches: \textbf{(1)} both the ResNet50 and simple CNN encoders were used in a supervised manner. In order to hold a fair comparison with the self-supervised approach, the same U-Net architecture as shown in Figure \ref{FIG:ssim_pipeline} was chosen for this, where both encoder and decoder were trained with randomly initialised weights. Approach \textbf{(2)} employs the identical encoders, however, only the decoder is trained; both encoder and decoder weights are still randomly initialised. In other words, the encoder weights are frozen and were not updated during training (i.e. back-propagation).

All other parameters and settings are kept the same for all approaches and identical to the SSL approach for the semantic segmentation task. In essence, the main task for the SSL approach is almost identical to the supervised approach, with the exception being that in case \textbf{(1)} the encoder is also trained.

\section{Experimental Design}\label{sec:design}

\subsection{Data Distribution}\label{sec:data_distribution}

\begin{figure}[htb!]
\centering
\includegraphics[width=0.5\textwidth]{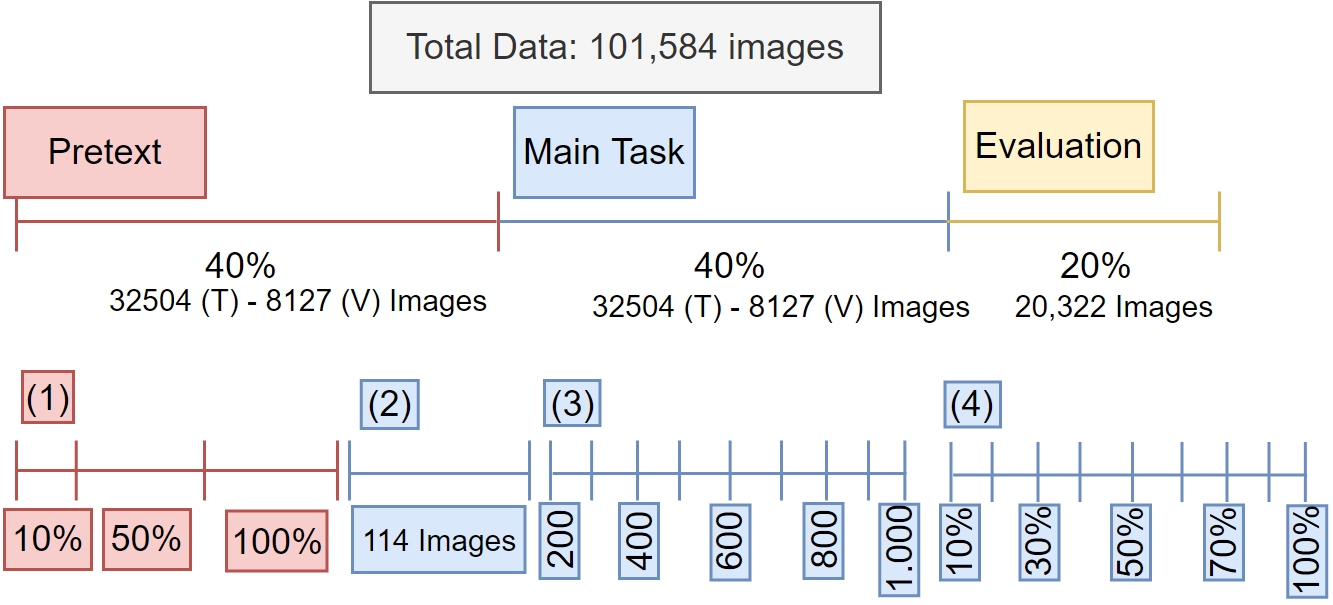}
    \caption{Visual of the distribution of the data, with three different training scenarios. (1) indicates the distribution only for the pretext task. (2), (3) and (4) indicate the data for the main task used in both SSL and supervised approaches. (T) denotes training while (V) denotes validation.
    \label{FIG:data_dist}}
\end{figure}

The data set, as described in section \ref{subsec:data}, was shuffled and randomly divided into three subsections in an equal distribution from each CZI file: $Train_{pretext}$, $Train_{main}$, and $Evaluation$. As shown in the top half of Figure \ref{FIG:data_dist}, from the roughly 100.000 crops, 40\% was reserved for $Train_{pretext}$, 40\% was taken for $Train_{main}$, and the remaining 20\% for $Evaluation$. 

The data set was separated in this manner prior to any training as to ensure that at each stage, the model is trained on images it has never seen before; this is done to prevent over-fitting. Additionally, five-fold cross-validation was used to ensure stability during the training of each architecture.
Four training scenarios, illustrated in Figure \ref{FIG:data_dist} and Table \ref{tab:case_summary}, were define to evaluate and compare supervised and self-supervised frameworks: \textbf{(1)} the self-supervised framework is pre-trained on the $Train_{pretext}$ data set, which was further sub-divided into another three categories shown in table \ref{tab:pretext_task_data}, as well as the red illustrations in Figure \ref{FIG:data_dist}. This subdivision of the data set is done to observe the importance of pre-training the networks prior to transferring the knowledge and to assess the performance of the models on a various number of images. In this stage, the performance of the augmentation techniques and loss functions will also be measured. In order to evaluate the performance in this regard, the model is then further trained for the main task on 114 images taken from $Train_{main}$ and evaluated on the $Evaluation$ set.
\textbf{(2)} Both supervised and self-supervised are trained on the $Train_{main}$ data set, which again was subdivided. In this case, to observe the self-supervised method's ability to accurately segment images given a small number of labelled data, both supervised and self-supervised are trained on 114 images.
\textbf{(3)} To observe the point at which the supervised and self-supervised models have similar performances, both networks were also trained from 200 up to 1.000 images, with 100 image increments. 
\textbf{(4)} Lastly, the supervised model was trained on 10\% up to 100\% of the $Train_{main}$ data with 10\% increments to observe performances with large data sets. 
It is important to note that the same images were used for all training scenarios to establish a fair analysis. 
A short summary of these four experimental cases is displayed in Table \ref{tab:case_summary}.\\

\begin{table}[!ht]
 \centering
 \caption{The number of crops used for training the pretext tasks. The first column lists the percentage of the total images considered, out of which the number of the images used to train and validate the model are described in the remaining columns.}
\begin{tabular}{lcc}
  \toprule
   Percentage of data & Train set & Validate set \\
  \midrule
  10\% & 3250 &  813 \\
  \hline
  50\% & 16252 &  4063 \\
  \hline
  100\% & 32504 &  8127 \\
  \bottomrule
  
\end{tabular}
\label{tab:pretext_task_data}
\end{table}

\begin{table}[!ht]
    \centering
    \caption{A short summary of the four experiment cases.}
    \begin{tabularx}{\linewidth}{X}
    Four Experiment Cases \\
    \hline
        \textbf{(1)} SSL framework trained on $Train_{pretext}$, then 114 images from $Train_{main}$, and lastly $Evaluation$ to observe performance of augmentation techniques, loss, and percentage of pretext training data.\\
    
    \hline
        \textbf{(2)} SSL and Supervised frameworks are compared with a minimal number (114) training images taken from $Train_{main}$. The models are evaluated on $Evaluation$.\\
        
    \hline
       \textbf{(3)} SSL and Supervised frameworks are compared by training from 200-1000 images (from $Train_{main}$) and evaluated on $Evaluation$ to observe at which point the frameworks have similar performances. \\
     \hline 
       
       \textbf{(4)} Supervised framework is trained from 10\% to 100\% of $Train_{main}$ with 10\% increments then evaluated on $Evaluation$, and compared to the SSL to observe supervised performances on large data sets.
    \end{tabularx}
    \label{tab:case_summary}
\end{table}

\subsection{Implementation Details}\label{sec:imp_details}
Regarding the model, the ResNet50 encoder and decoder architecture follows the same structure as discussed in \cite{zhang2018road}. The ResNet50 encoder building block consists of a dual of batch normalization followed by a ReLU activation and a 3x3 convolution. The building block's output is added to the input using the identity mapping function as illustrated in \cite{zhang2018road}. This block is repeated four times for the encoding component and another four for decoding only with the addition of an upsampling layer between each block. Furthermore, a skip connection is formed between the encoder and decoder between each block. The convolutional encoder block starts with a 320x320 input matching the cropped image size, mentioned in section \ref{subsec:data}. Throughout the four convolutional layers, the input size is halved. Hence, the second layer has a size of 160x160, the third 80x80, and the last layer has a size of 40x40 pixels. The decoder performs this in the reversed order.

During the training phase for both the pretext and the main task, the model was trained over 50 epochs. The Adam optimizer was used as the learning scheduler with a learning rate of $0.003$. The data was divided into a batch size of 16, and a seed value of 26 was used to ensure reproducibility when using random variables (i.e. shuffling the batches). Regarding constants used in the loss functions, the $L_{SSIM}$ and $L_{SSIM-L1}$ functions had $c_1 = 0.01$ and $c_2 = 0.03$, whilst $L_{dice}$ and $L_{IoU}$ had $\epsilon = 0.0001$. Lastly, one Nvidia V100 GPU accelerator card was used for training all models.
\section{Results}\label{sec:results}
Due to the nature of the pixel-wise binary classification task a confusion matrix was computed for each image that has been segmented. From this, the accuracy, precision, recall, F1-score, and Jaccard index was computed. The metric that we are most interested in is the \textbf{F1-Score}, in some cases called the harmonic mean. The F1-score penalizes large differences between precision and recall, which sets apart the desirable image segmentations from the undesirable ones.

\begin{figure*}[ht!]
\centering
\includegraphics[width=1\textwidth]{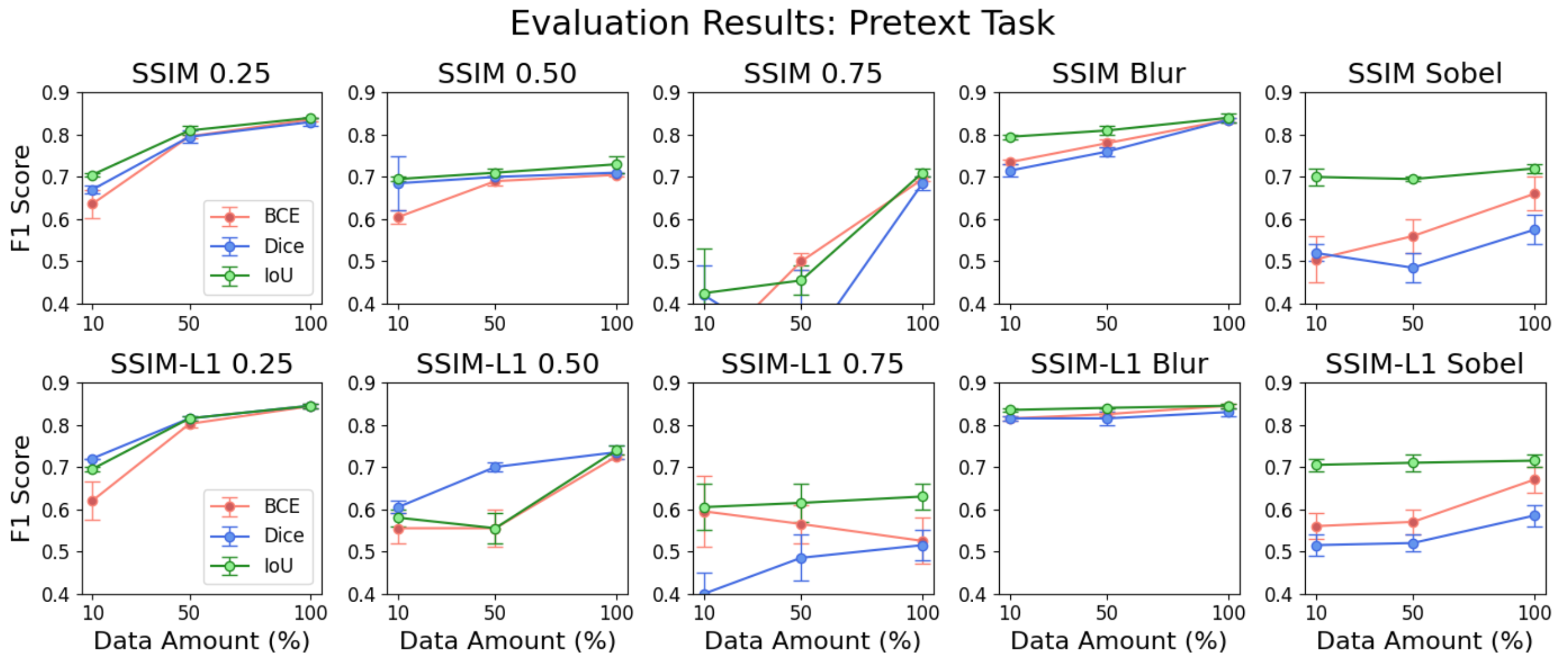}
    \caption {Evaluation of the self-supervised pretext tasks, described in section \ref{subsec:data}, using F1-scores. The mean of the 5 folds for the BCE, Dice, and IoU loss functions were computed and is indicated by the red, blue, and green points, respectively. The top row displays the scores using the SSIM loss of the pretext task, while the bottom row displays the SSIM-L1 loss. }
    \label{fig:eval_pretext}
\end{figure*}

\subsection{Self-Supervised Framework}\label{sec:self-supervised_results}

\begin{table*}[ht!]
\caption{
Self-Supervised architecture, F1-Scores, on main task after training on reconstruction (25\% pixel drop) pretext task, 114 images.
The best score, average score and the standard deviation of the five folds cross validation is computed for each variation of network structure.}
\label{tab:f1_scores_self-sup_pd_avg}
\centering
\resizebox{\textwidth}{!}{%
\begin{tabular}{ll|cccccc|cccccc}
    \hline 
       && \multicolumn{12}{c}{Self-Supervised (0.25 pixel drop)}\\
    \hline
    &  Pre-Train & \multicolumn{6}{c}{SSIM} & \multicolumn{6}{c}{SSIM-L1} \\
    \hline
    \multirow{5}{*}{\rotatebox[origin=c]{90}{F1-Score}} &  & \multicolumn{2}{c}{BCE} & \multicolumn{2}{c}{Dice} & \multicolumn{2}{c|}{IoU} & \multicolumn{2}{c}{BCE} & \multicolumn{2}{c}{Dice} & \multicolumn{2}{c}{IoU} \\
     &  & Best & Mean $\pm$ Std & Best & Mean $\pm$ Std & Best & Mean $\pm$ Std & Best & Mean $\pm$ Std & Best & Mean $\pm$ Std & Best & Mean $\pm$ Std  \\
    & 10\% & 0.69 & 0.64 $\pm$ 0.034  & 0.66 &  0.64 $\pm$ 0.028 & 0.71 & 0.70 $\pm$ 0.005 & 0.58 & 0.54 $\pm$ 0.047 & 0.71 &
        0.70 $\pm$ 0.04 & 0.70 & 0.69 $\pm$ 0.02\\
    & 50\% & 0.80 & 0.80 $\pm$ 0.005 & 0.81 & 0.78 $\pm$ 0.024 & 0.81 & 0.80 $\pm$ 0.012 & 0.81 & 0.80 $\pm$ 0.012 & 0.82 &
        0.81 $\pm$ 0.005 & 0.82 & 0.82 $\pm$ 0.007\\
    & 100\% & 0.84 & 0.84 $\pm$ 0.005 & 0.84 & 0.82 $\pm$ 0.012 & 0.84 & 0.84 $\pm$ 0.005 & 0.85  & 0.85 $\pm$ 0.005 & 0.84 & 0.84 $\pm$ 0.004 & 0.85 & 0.85 $\pm$ 0.005\\
    \hline
\end{tabular}
}
\end{table*}

\begin{table*}[ht!]
\caption{
Self-Supervised architecture, F1-Scores, on main task after training on deblurring pretext task, 114 images.
The best score, average score and the standard deviation of the five folds cross validation is computed for each variation of network structure.}
\label{tab:f1_scores_self-sup_blur_avg}
\centering
\resizebox{\textwidth}{!}{%
\begin{tabular}{ll|cccccc|cccccc}
    \hline 
       && \multicolumn{12}{c}{Self-Supervised (Blurring)}\\
    \hline
    &  Pre-Train & \multicolumn{6}{c}{SSIM} & \multicolumn{6}{c}{SSIM-L1} \\
    \hline
    \multirow{5}{*}{\rotatebox[origin=c]{90}{F1-Score}} &  & \multicolumn{2}{c}{BCE} & \multicolumn{2}{c}{Dice} & \multicolumn{2}{c|}{IoU} & \multicolumn{2}{c}{BCE} & \multicolumn{2}{c}{Dice} & \multicolumn{2}{c}{IoU} \\
     &  & Best & Mean $\pm$ Std & Best & Mean $\pm$ Std & Best & Mean $\pm$ Std & Best & Mean $\pm$ Std & Best & Mean $\pm$ Std & Best & Mean $\pm$ Std  \\
    & 10\% & 0.74 & 0.73 $\pm$ 0.015 & 0.73  & 0.70 $\pm$ 0.014 & 0.80 & 0.79 $\pm$ 0.014 & 0.82 & 0.81 $\pm$ 0.010 & 0.82
    & 0.81 $\pm$ 0.08 & 0.84 & 0.84 $\pm$ 0.005\\
    & 50\% & 0.79 & 0.77 $\pm$ 0.01 & 0.77 & 0.75 $\pm$ 0.19 & 0.82 & 0.81 $\pm$ 0.008 & 0.83 & 0.81 $\pm$ 0.008 & 0.83 & 0.82 $\pm$ 0.012 & 0.84 & 0.84 $\pm$ 0.005 \\
    & 100\% & 0.84 & 0.84 $\pm$ 0.005 & 0.84 & 0.84 $\pm$ 0.005 & 0.85 & 0.84 $\pm$ 0.005 & 0.85 & 0.85 $\pm$ 0.005 & 0.84 & 0.84 $\pm$ 0.005 & 0.85 & 0.85 $\pm$ 0.000\\
    \hline
\end{tabular}
}
\end{table*}



\begin{table*}[ht!]
\caption{Supervised architecture F1-Scores trained on 114 images. The best score, average score and the standard deviation of the five folds cross validation is computed for each variation of network structure.}
\label{tab:f1_scores_supervised_avg}
\centering
\resizebox{\textwidth}{!}{%
\begin{tabular}{ll|cccccc|cccccc}
    \hline 
       && \multicolumn{12}{c}{Supervised}\\
    \hline
    &  Pre-Train & \multicolumn{6}{c}{Freeze} & \multicolumn{6}{c}{No-Freeze} \\
    \hline
    \multirow{5}{*}{\rotatebox[origin=c]{90}{F1-Score}} &  & \multicolumn{2}{c}{BCE} & \multicolumn{2}{c}{Dice} & \multicolumn{2}{c|}{IoU} & \multicolumn{2}{c}{BCE} & \multicolumn{2}{c}{Dice} & \multicolumn{2}{c}{IoU} \\
     &  & Best & Mean $\pm$ Std & Best & Mean $\pm$ Std & Best & Mean $\pm$ Std & Best & Mean $\pm$ Std & Best & Mean $\pm$ Std & Best & Mean $\pm$ Std  \\
    & & & & & & & \\
    & ResNet50 & 0.75 & 0.46 $\pm$ 0.232 & 0.60 & 0.22 $\pm$ 0.199 & 0.47 & 0.32 $\pm$ 0.137 & 0.57 & 0.43 $\pm$ 0.118 & 0.73 & 0.49 $\pm$ 0.254 & 0.78 & 0.644 $\pm$ 0.08 \\
    & CNN & 0.63 & 0.37 $\pm$ 0.161 & 0.27 & 0.054 $\pm$ 0.108 & 0.27 & 0.262 $\pm$ 0.016 & 0.71 & 0.44 $\pm$ 0.195 & 0.27 & 0.11 $\pm$ 0.132 & 0.27 & 0.27 $\pm$ 0.000 \\
    \hline
    
\end{tabular}
}
\end{table*}

Figure \ref{fig:eval_pretext} displays the results produced by the SSL framework first trained on $Train_{pretext}$, then on 114 images from $Train_{main}$, using the SSIM and SSIM-L1 for the pretext, then BCE, Dice, and IoU for the main task. It can be observed that, generally, as the percentage of $Train_{pretext}$ data increases, the scores across all metrics also increase for all three main task loss functions. This confirms that increasing the amount of pre-trained data will positively influence the main task.

Another observation that can be made is that out of the five augmentation types, the 25\% pixel drop and the blurring methods produce the best results reaching an F1-score as high as 0.85, which is considerably higher than the other three techniques where an F1-score of 0.75 was the highest. This would suggest that changing the image too strongly will make it more difficult for the network to restore the images and extract the mid to low-level features. This is especially the case for the 75\% pixel drop augmentation, which scored the lowest at 0.25 for the SSIM loss with 10\% $Train_{pretext}$ and 0.4 for the SSIM-L1. Augmenting using Sobel filtering shows better but similar results. In both cases, this relates to how both these augmentation techniques cause the pixels to be affected the most. Despite this, scores can still be improved as a general trend by increasing the amount of training data for the pretext task. 

Regarding loss functions, it can be observed that the IoU loss, indicated by the green points, achieved the highest scores reaching 0.85 when trained with 100\% of $Train_{pretext}$ using either SSIM or SSIM-L1 for the pretext task. Furthermore, the IoU loss, on average, performed best regardless of augmentation technique, with only the SSIM-L1 using the 50\% pixel drop being the outlier. This confirms that adding scale invariance (section \ref{sec:iou}) is effective for the organoids data set due to the smaller organoids being scattered across each image.

Figures \ref{fig:segmentations_10}, \ref{fig:segmentations_50}, and \ref{fig:segmentations_100} in the supplementary illustrate the segmentation masks generated for the various loss functions in combination with the augmentation techniques using the SSIM loss. The Figures visually indicate how well the 25\% pixel drop and Gaussian blurring has performed when compared to the other augmentation techniques. Furthermore, it can be observed that as the percentage of pretext training data increases, the generated segmentation masks come closer to the true masks regardless of augmentation technique, once again confirming its influence over the main task. The SSIM-L1 loss has generated similar masks, which demonstrates a similar trend and was therefore left out in this case. 






 


\subsection{Comparing Frameworks}
 Tables \ref{tab:f1_scores_self-sup_pd_avg}, \ref{tab:f1_scores_self-sup_blur_avg}, and \ref{tab:f1_scores_supervised_avg} display the best F1-scores, average F1-score as well as the deviations for both self-supervised and supervised frameworks, trained on 114 images.
 The deviations can be used as an indicator for model stability, with larger deviations meaning lower stability and smaller deviations for higher stability. For the self-supervised approach, only the scores for the 25\% pixel drop and blurring augmentation are displayed, as these two techniques reported the strongest results, as shown in section \ref{sec:self-supervised_results}.

When observing the results of the blurring technique, shown in Table \ref{tab:f1_scores_self-sup_blur_avg}, a particular point of interest is that the 10\% SSIM-L1 to IoU network performs just as well as the 100\% SSIM to BCE and Dice, all cases having an F1-score of 0.84. This confirms that suppressing colour changes and brightness with the addition of the L1 loss to the SSIM loss can be effective (mentioned in section \ref{sec:ssim_l1_loss}). The 25\% pixel drop, shown in Table \ref{tab:f1_scores_self-sup_pd_avg}, was able to achieve high F1-scores of up to 0.85 as well. However, when using 10\% of $Train_{pretext}$, the scores range from 0.58-0.71, while the blurring technique, in contrast, achieved a range of 0.73-0.84, which is substantially higher. 

For the supervised approach shown in Table \ref{tab:f1_scores_supervised_avg}, it can be seen that the ResNet50 encoder largely outperforms the simple CNN encoder in regards to the F1-scores, as shown by the highest score of 0.78 for the ResNet50 encoder compared to the 0.63 for the simple CNN encoder. This implies that the complexity of the encoder plays an important role in optimizing improvements. In a similar fashion to the self-supervised approach, the IoU loss appears to perform the best in the supervised context achieving a score of 0.78. Interestingly, ResNet50 with frozen encoders trained on BCE also performed strongly with a score of 0.75. Another point of interest in regards to the simple CNN architecture is that BCE seems to be the only loss function to effectively produce higher scores of 0.63 and 0.71, compared to 0.27 for the Dice and IoU losses. It also appears to be the case that scores do not differ too strongly when comparing the frozen and non-frozen encoders. This could suggest that the decoder does the majority of the work in semantic segmentation tasks, something that is also in agreement with \cite{goutam2020layerout}.

\begin{figure}[htb!]
\centering
\includegraphics[width=0.48\textwidth]{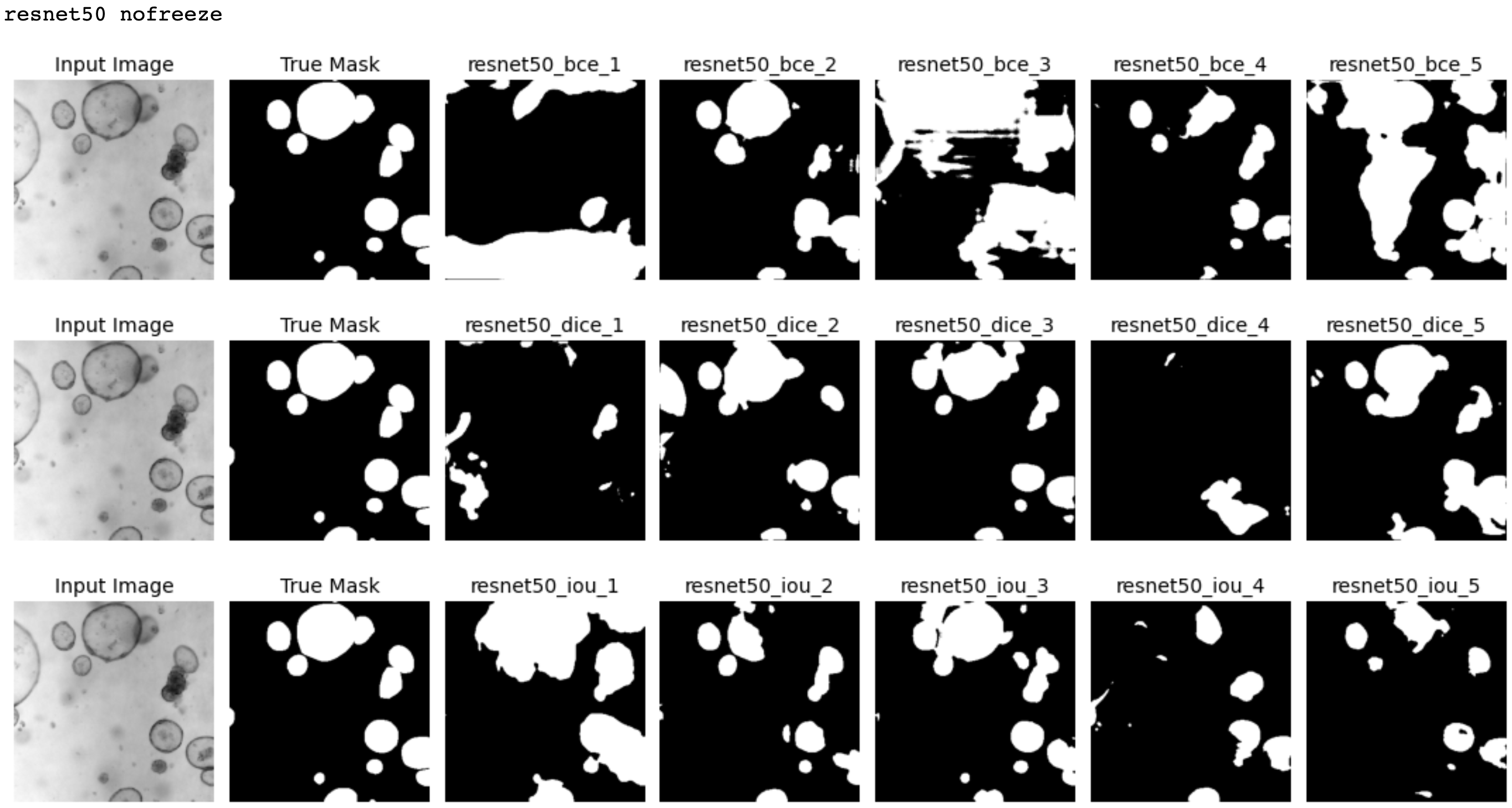}
    \caption {Segmentations made with the ResNet50 encoder using supervised approach, where the encoder weights \textit{are not} frozen. On display from the left is: Input (Ground Truth) image, True Mask, then the generated masks of the five folds in ascending order. The top row illustrates the masks using the BCE loss, the middle row using the Dice loss, and the bottom using the IoU loss.}
    \label{fig:segmentations_resnet_nofreeze}
\end{figure}

\begin{figure}[htb!]
\centering
\includegraphics[width=0.48\textwidth]{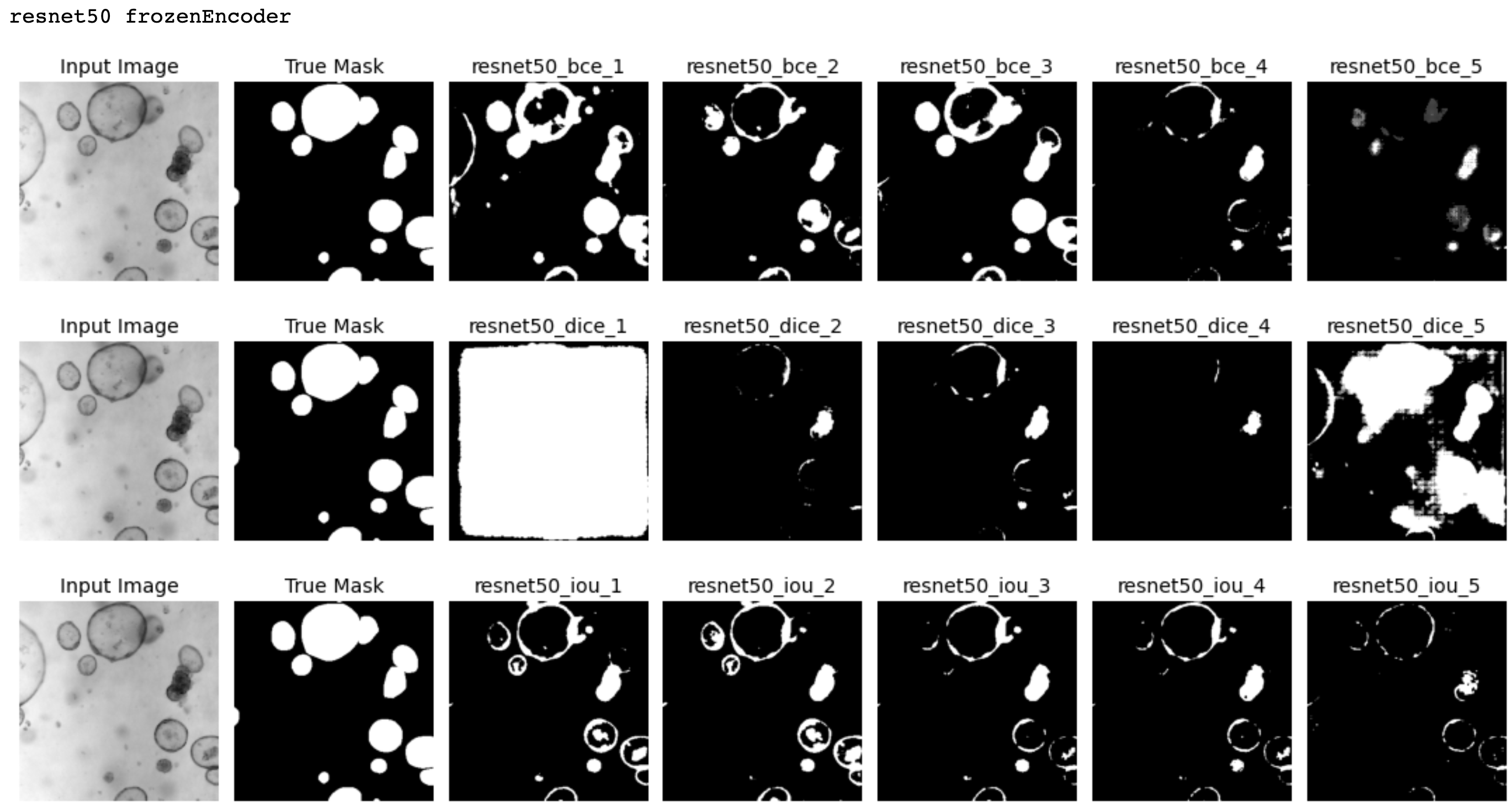}
    \caption {Segmentations made with the ResNet50 encoder using supervised approach, where the encoder weights \textit{are} frozen. On display from the left is: Input (Ground Truth) image, True Mask, then the generated masks of the five folds in ascending order. The top row illustrates the masks using the BCE loss, the middle row using the Dice loss, and the bottom using the IoU loss.}
    \label{fig:segmentations_resnet_frozen}
\end{figure}

\begin{figure}[htb!]
\centering
\includegraphics[width=0.48\textwidth]{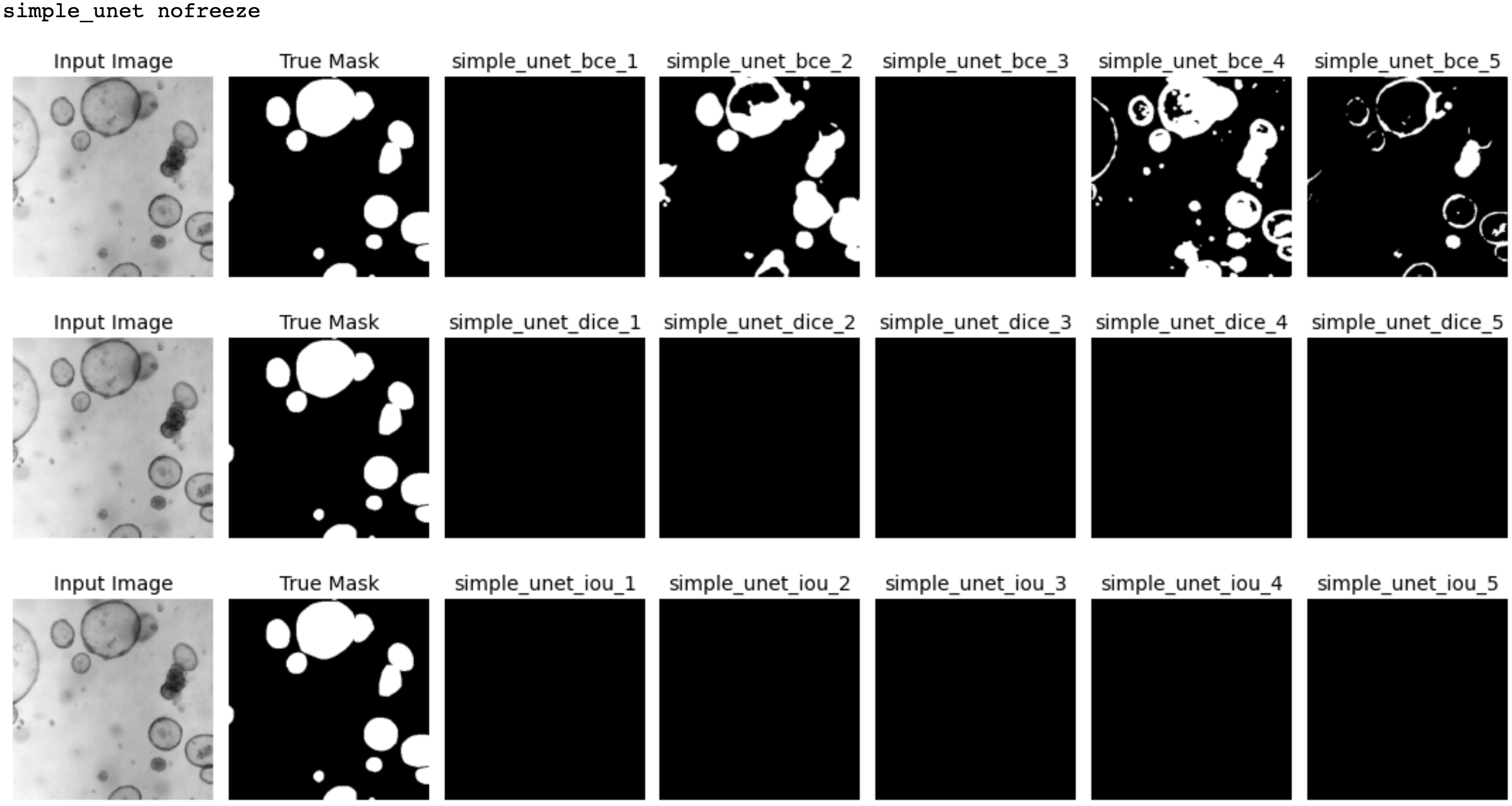}
    \caption {Segmentations made with the simple CNN encoder using supervised approach, where the encoder weights \textit{are not} frozen. On display from the left is: Input (Ground Truth) image, True Mask, then the generated masks of the five folds in ascending order. The top row illustrates the masks using the BCE loss, the middle row using the Dice loss, and the bottom using the IoU loss.}
    \label{fig:segmentations_cnn_nofreeze}
\end{figure}

\begin{figure}[htb!]
\centering
\includegraphics[width=0.48\textwidth]{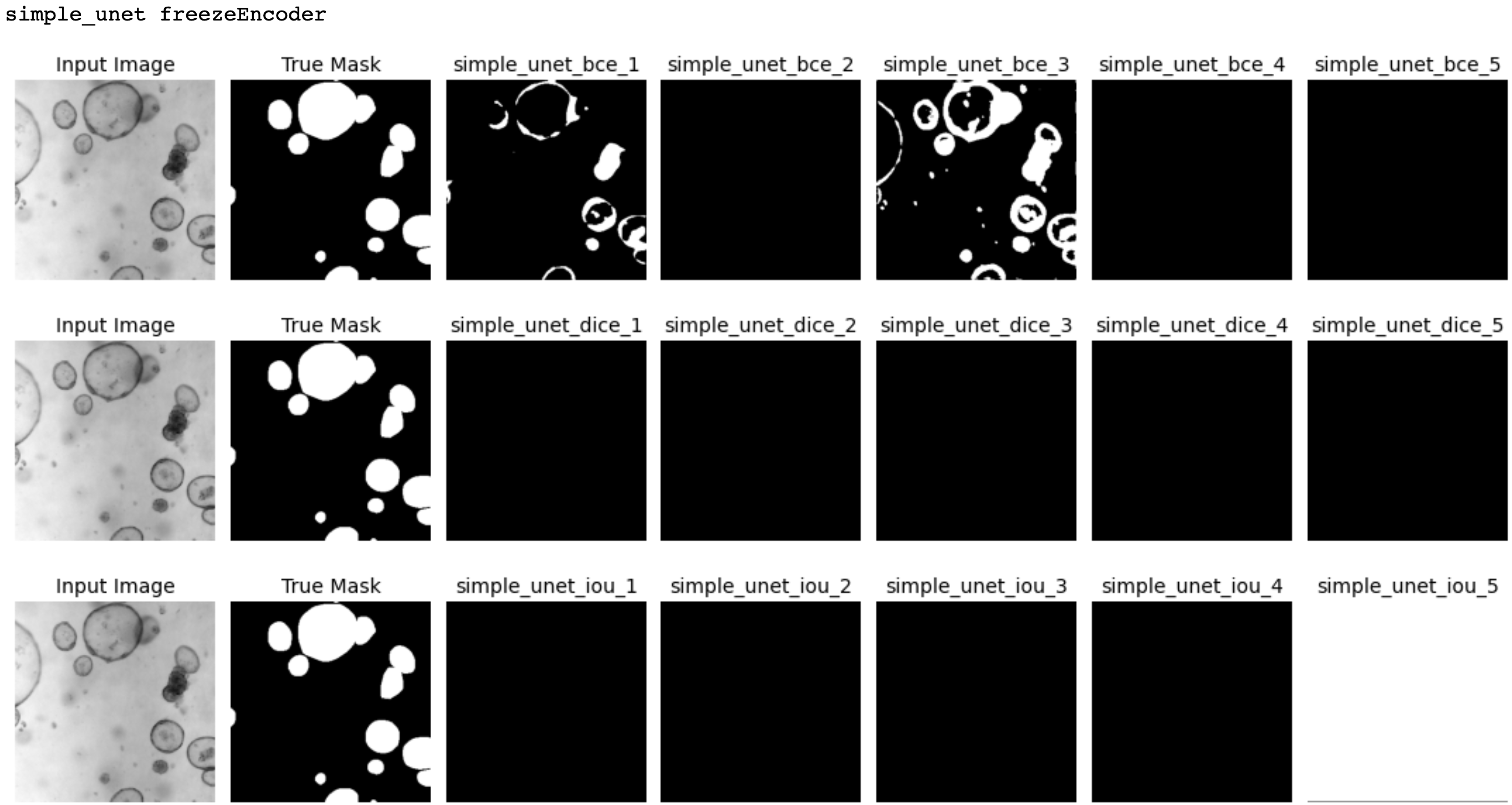}
    \caption {Segmentations made with the simple CNN encoder using supervised approach, where the encoder weights \textit{are} frozen. On display from the left is: Input (Ground Truth) image, True Mask, then the generated masks of the five folds in ascending order. The top row illustrates the masks using the BCE loss, the middle row using the Dice loss, and the bottom using the IoU loss.}
    \label{fig:segmentations_cnn_frozen}
\end{figure}

Figures \ref{fig:segmentations_resnet_nofreeze}, \ref{fig:segmentations_resnet_frozen}, \ref{fig:segmentations_cnn_nofreeze}, and \ref{fig:segmentations_cnn_frozen} illustrate the segmentations performed by both ResNet50 and CNN using supervised frameworks across the five-folds cross-validation models. Regarding the ResNet50 architecture, it can be observed that when the encoder \textit{is not} frozen, the generated segmentation masks are able to fill in the organoid shape. In contrast, when the encoder \textit{is} frozen, it is only able to segment the edges. As for the simple CNN encoder, it is rather clear here that the model is unable to converge in most cases, with the exception of the BCE loss, which, as discussed earlier, was the only loss function to produce meaningful results. This could be due to the limited amount of data (114 images) that is available for training. 

Lastly, when comparing the two frameworks, it can be observed that with a small data set (114 images), the SSL framework performs better than either ResNet50 or a simple CNN using the supervised framework. For instance, the F1-score for the blur augmentation technique (Table \ref{tab:f1_scores_self-sup_blur_avg}) was between 0.73-0.85, which was generally higher than the supervised framework (Table \ref{tab:f1_scores_supervised_avg}) having best scores between 0.27-0.78. Additionally, the self-supervised approach is able to perform well consistently regardless of the loss function. The supervised approach, in contrast, has strong inconsistencies in this aspect. Furthermore, when observing the deviations in Tables \ref{tab:f1_scores_self-sup_pd_avg}, \ref{tab:f1_scores_self-sup_blur_avg}, and \ref{tab:f1_scores_supervised_avg}, a clear disparity can be observed in stability between the SSL and the supervised approach, where the SSL has at most a deviation of 0.047 whilst the supervised has a deviation as high as 0.254.

\begin{figure}[htb!]
\centering
\includegraphics[width=0.48\textwidth]{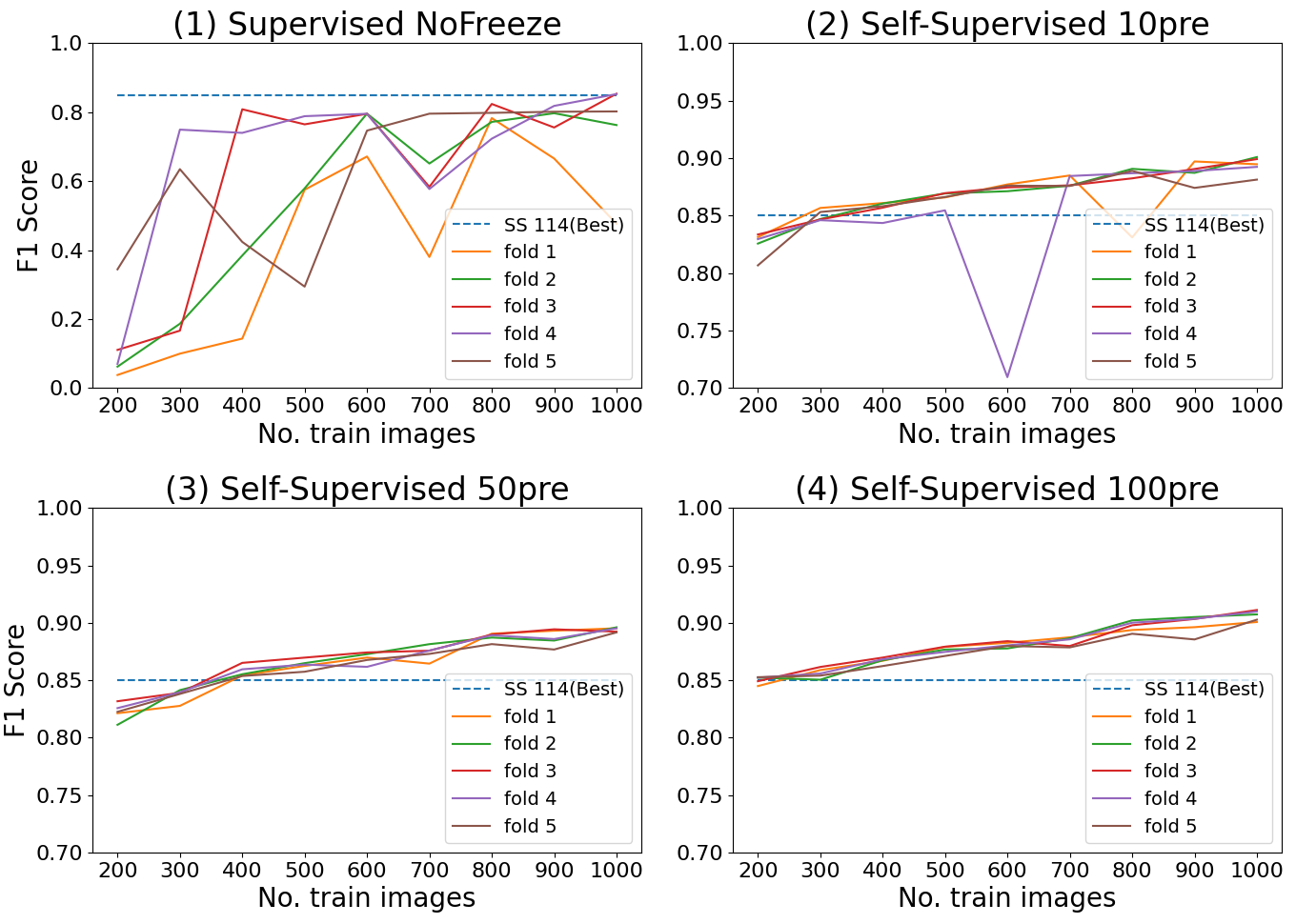}
    \caption {F1-scores for (1) supervised trained with the IoU loss using the frozen ResNet50 encoder, and (2)(3)(4) self-supervised trained between 200 up to 1000 images using the SSIM-L1 and IoU loss with 10\%, 50\%, and 100\% of $Train_{pretext}$ data. The blue dotted line indicates the best F1-score of the self-supervised approach trained on 114 images, and is used as a baseline benchmark. Each full line indicates an individual model in the 5 fold cross validation. SSL is trained using blurred crops.}
    \label{fig:comparative_perf}
\end{figure}

\begin{figure*}[htb!]
\centering
\includegraphics[width=1\textwidth]{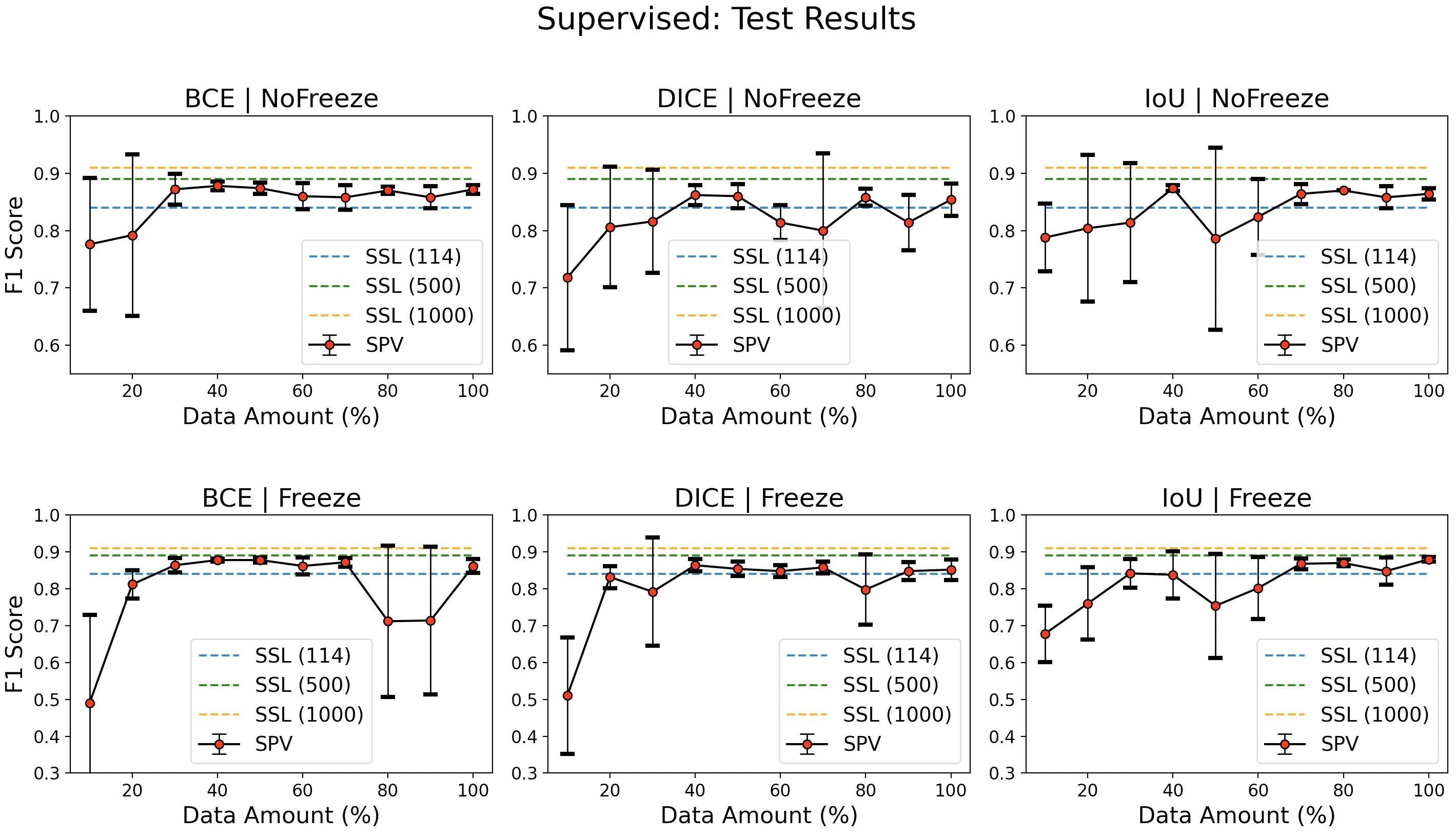}
    \caption {Evaluation of supervised (SPV) using F1-Scores trained on various percentage of $Train_{main}$. The mean of the 5 folds was computed and is indicated by the red points, with bars denoting the standard deviation. This is compared to the mean of the self-supervised (SSL) scores shown by the dotted lines. The SSL network was trained with the SSIM-L1 to IoU losses with: 114, 500, and 1.000 images. The top row shows the scores when the encoder is not frozen, whilst the bottom row shows the frozen encoder.}
    \label{fig:supervised_test_results}
\end{figure*}

\subsection{Comparing Performance with Larger Data Sets}
To further observe performance differences between the supervised and the self-supervised approach, the F1-score was measured across networks trained with various numbers of images. Figure \ref{fig:comparative_perf} displays the F1-scores when trained between 200-1.000 images for all five CV folds. From graph \textbf{(1)}, illustrating the standard supervised approach with a ResNet50 encoder without frozen weights, it can be observed that it takes upwards of 1.000 training images to achieve similar results to the best self-supervised approach that was trained on 114 images.

Graphs \textbf{(2)}-\textbf{(4)}, display the scores for the self-supervised approach pre-trained on the deblurring pretext task with 10\%, 50\%, and 100\% of the sub-data, respectively (Table \ref{tab:pretext_task_data}). A major point of interest here is that F1 scores improve when trained with more data on the main task while remaining ahead of the supervised approach. Additionally, when comparing the curves of both approaches, it is clear here that the self-supervised approach has improved stability and reliability over the supervised one. This is a strong indicator of the effectiveness of the self-supervised approach.

\subsection{Supervised Framework}

Figure \ref{fig:supervised_test_results} displays the F1-scores for the supervised (SPV) framework trained between 10\%-100\% of $Train_{main}$ that is compared to the self-supervised (SSL) framework trained with 114, 500, and 1.000 images. As a first observation, the SPV scores outperform the SSL-114 scores when trained with 30\%  of $Train_{main}$ or more, but not the SSL-500 and SSL-1000. When comparing the frozen and non-frozen encoders, there appears to be little difference in scores which, again, is in agreement with \cite{goutam2020layerout} regarding the difference in effectiveness between the encoder and decoder. Lastly, the deviation bars indicate the stability of training across the 5-fold cross-validation, where BCE appears to have the highest stability.

\section{Conclusions}\label{sec:conclusions}

\subsection{Discussion}
This work evaluates the potential of using the SSL paradigm to perform semantic segmentation on images of organoids, specifically by employing the U-Net architecture through image restoration as a pretext task. The use of SSL methods in the context of medical has the potential to greatly improve research in the field by automating processes that otherwise require manual labour from experts. 

Regarding the proposed pretext task, a total of 5 augmentation techniques (25\%, 50\%, 75\% pixel drops, Gaussian blurring, and Sobel filtering) paired with 2 loss functions (SSIM and SSIM-L1), trained across 10\%, 50\% and 100\% of data taken from $Train_{pretext}$ were compared. The gained knowledge from the pretext task was then transferred to the main task where the BCE, Dice, and IoU losses were compared afterwards, being trained on 114 images taken from $Train_{main}$. Here, it was discovered that the Guassian blurring augmentation paired with the SSIM-L1 and IoU losses achieved the best results across the range of $Train_{pretext}$ data, with F1-scores of 0.84-0.85, with the 25\% pixel drop (with the same losses) coming in close second with a range of 0.70-0.85. Furthermore, it can be concluded that across all metrics in this context, the IoU loss performed the best.

When comparing the SSL and supervised frameworks, it is abundantly clear here that the self-supervised approach is able to achieve better scores than the supervised with smaller labelled data sets. When trained on 114 images from $Train_{main}$, the SSL framework was able to achieve an F1-score of 0.85, whilst the supervised framework scored at most 0.78. It is also essential to emphasise the importance of using a complex encoder such as the ResNet50 compared to a CNN, as shown by the comparison between the two supervised frameworks. With larger data sets, a similar conclusion can be drawn that the SSL approach still outperforms the supervised one. 

An inherent deficiency in the SSL approach however, is that it will always take longer to train. In the case that labelled data is already available, it would be more efficient and time effective to take the supervised approach. As mentioned previously however, this in general is not the case with medical imaging. Regardless, it can safely be concluded here that the SSL paradigm can be used effectively to produce robust models when labelled data is not available.

\subsection{Future Works}
A potential for further investigation regarding the topic of semantic segmentation of organoids could be the use of a different encoder architectures such as the VGG16 \cite{simonyan2014very} or the MobileNet \cite{howard2017mobilenets} architectures which have been proven to perform strongly in computer vision tasks. Considering that the decoder plays a significant role in the task of segmentation, it would be beneficial to study the potential of other decoders in this context. Another point of interest could be the use of a different pretext task in the self-supervised method, as this also plays an important role in improving the overall performance. On the topic of pretext task, it would also be interesting to observe the performance of the image reconstruction task on other data sets such as the COCO \cite{lin2014microsoft} set.

\printcredits

\bibliographystyle{cas-model2-names}

\bibliography{cas-refs}


\clearpage
\section*{Supplementary Materials}\label{sec:supplementary}


\begin{figure}[htb!]
\centering
\includegraphics[width=0.48\textwidth]{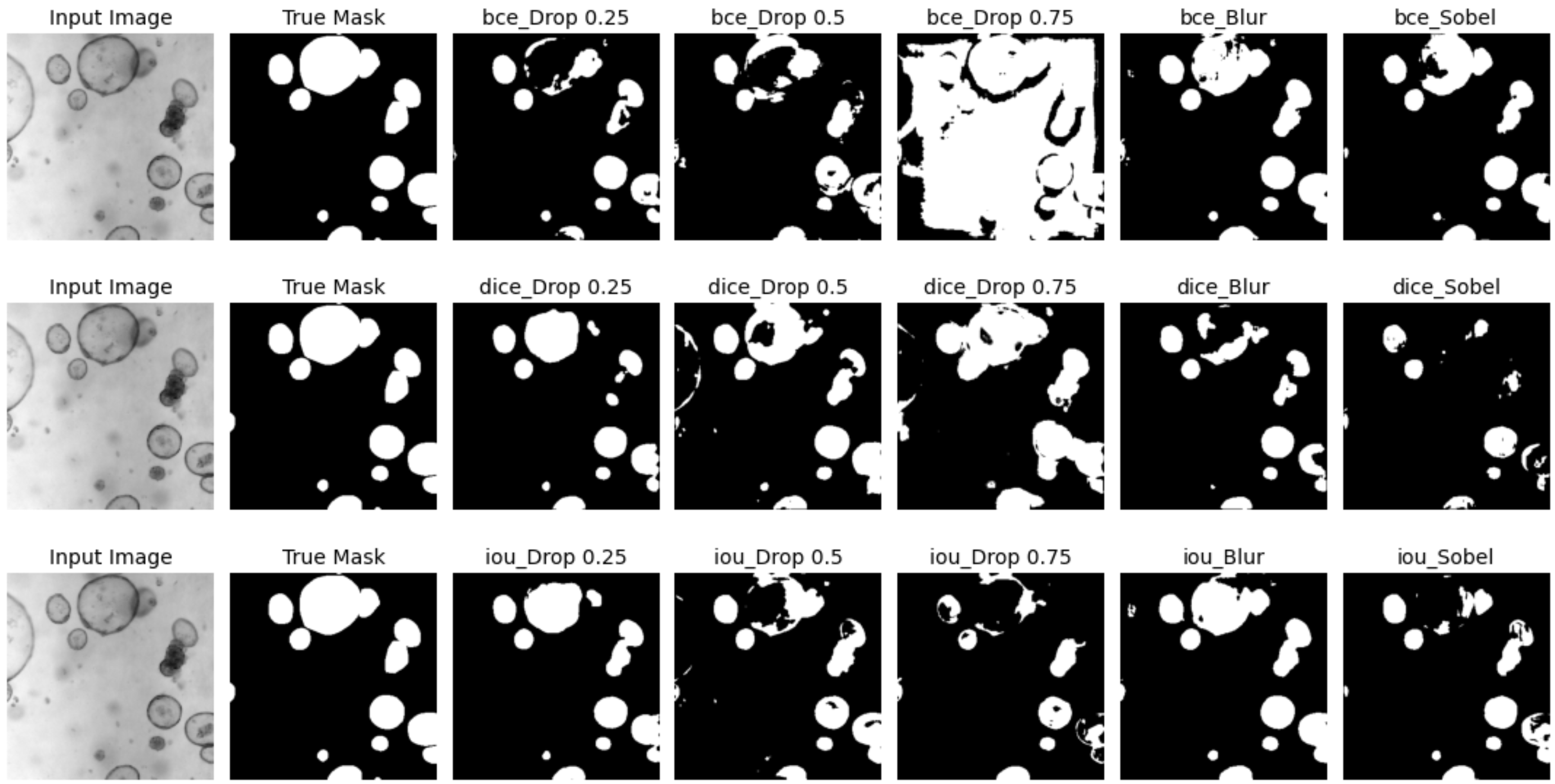}
    \caption {Segmentation masks generated by the SSL framework using 10\% of $Train_{pretext}$. From left to right is: Input (Ground Truth) Image, True Mask, 25\%, 50\%, 75\% Pixel Drop, Gaussian Blurring, Sobel Filtering. The top row illustrates the masks using the BCE loss, the middle row using the Dice loss, and the bottom using the IoU loss.}
    \label{fig:segmentations_10}
\end{figure}

\begin{figure}[htb!]
\centering
\includegraphics[width=0.48\textwidth]{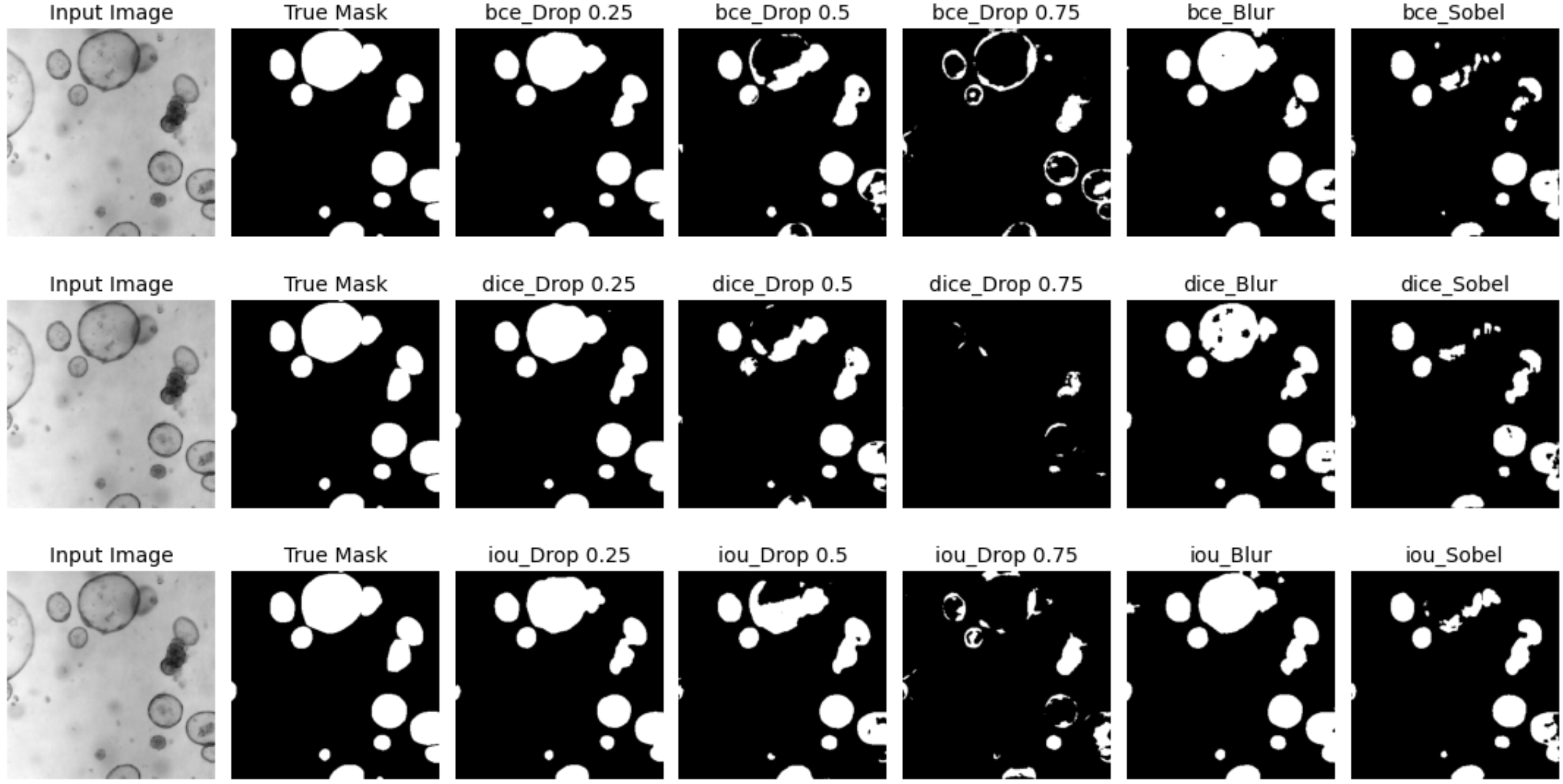}
    \caption {Segmentation masks generated by the SSL framework using 50\% of $Train_{pretext}$. From left to right is: Input (Ground Truth) Image, True Mask, 25\%, 50\%, 75\% Pixel Drop, Gaussian Blurring, Sobel Filtering. The top row illustrates the masks using the BCE loss, the middle row using the Dice loss, and the bottom using the IoU loss.}
    \label{fig:segmentations_50}
\end{figure}

\begin{figure}[htb!]
\centering
\includegraphics[width=0.48\textwidth]{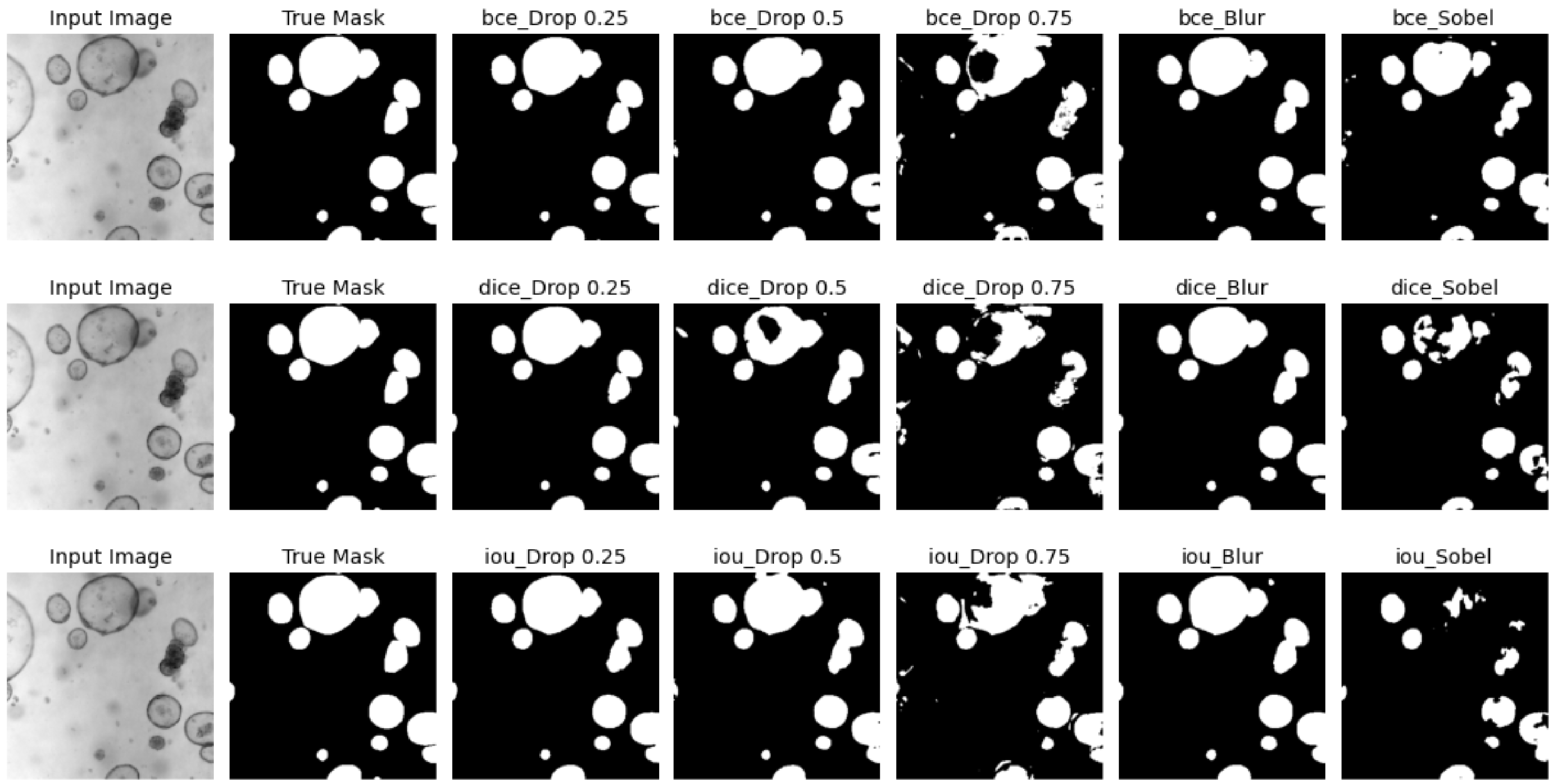}
    \caption {Segmentation masks generated by the SSL framework using 100\% of $Train_{pretext}$. From left to right is: Input (Ground Truth) Image, True Mask, 25\%, 50\%, 75\% Pixel Drop, Gaussian Blurring, Sobel Filtering. The top row illustrates the masks using the BCE loss, the middle row using the Dice loss, and the bottom using the IoU loss.}
    \label{fig:segmentations_100}
\end{figure}

\clearpage


\end{document}